\documentclass{article}

\usepackage[preprint]{corl_2026} 

\usepackage{amsmath}
\usepackage{amssymb}
\usepackage{graphicx}
\usepackage{caption}
\usepackage{float}

\title{SurgVIL: Scaling Surgical Robot Imitation \\ Learning with Open-source Surgical Videos}

\author{
  Xinhao Chen\\
  Laboratory for Computational Sensing and Robotics (LCSR)\\
  Johns Hopkins University, 
  USA\\
  \texttt{xchen254@jhu.edu} \\
  \And
  JuoTung Chen\\
  LCSR\\
  Johns Hopkins University, 
  USA\\
  \And
  Nigel Nelson\\
  Nvidia, 
  USA\\
  \And
  Antony Goldenberg\\
  LCSR\\
  Johns Hopkins University, 
  USA\\
  \And
  Jesse Haworth\\
  LCSR\\
  Johns Hopkins University, 
  USA\\
  \And
  Sean D. Huver\\
  Nvidia, 
  USA\\
  \And
  Axel Krieger\\
  LCSR\\
  Johns Hopkins University, 
  USA\\
}

\begin{document}
\maketitle


\begin{abstract}
    Learning-based surgical robot autonomy requires large-scale demonstrations with synchronized videos and robot actions, but such data are exceedingly rare in clinical or realistic tissue settings because robot kinematics are typically inaccessible outside controlled research systems. In contrast, phantom data collected on research platforms provide accurate action labels but lack the visual diversity of real tissue. We propose SurgVIL, a framework for scaling surgical robot imitation learning using open-source surgical videos. SurgVIL combines kinematically labeled phantom robot demonstrations with surgical videos from open-source datasets and online sources for policy learning. Since these videos lack robot motion labels, we estimate approximate kinematics as weak supervision. We evaluate SurgVIL on two da Vinci robot tasks: needle pick-up and cholecystectomy cutting. Across ACT, $\pi 0$, and Gr00t-H backbones, adding surgical videos substantially improves generalization to real-tissue and out-of-distribution settings, suggesting a scalable path from phantom training toward generalizable surgical robot policies.
\end{abstract}


	
\section{Introduction}

Robotic surgical autonomy has the potential to improve the consistency, precision, and accessibility of robot-assisted procedures~\cite{attanasio2021autonomy}. Recent progress in surgical robotics and artificial intelligence has motivated increasing interest in autonomous or semi-autonomous surgical systems, ranging from perception assistance to task-level robotic execution~\cite{schmidgall2025will}. In parallel, imitation learning and vision-language-action models have shown strong promise for learning complex robot manipulation skills from demonstrations~\cite{kim2025srt, nelson2026open}. However, learning-based surgical autonomy remains strongly limited by the availability of large-scale, diverse, and accurately labeled surgical robot data.

For surgical robot policy learning, the most valuable data consist of synchronized endoscopic video and robot kinematics. Such data can be collected in controlled laboratory environments using phantoms, or in simulation environments, where robot kinematics can be recorded directly from the robot system or simulator. However, collecting the same type of synchronized robot-action data in realistic surgical settings is substantially more difficult. Collecting kinematic data from clinical robotic surgery in hospital is often restricted by regulatory constraints and by robot manufacturers. As a result, accurately labeled surgical robot datasets with clinically realistic visual diversity remain scarce.
Phantom demonstrations provide reliable action labels, but they alone are often insufficient for robust generalization. A policy trained only on phantom data may perform well when evaluated on the same phantom setup, yet fail when the visual background, tissue appearance, or target configuration changes. This limitation is particularly important for surgical manipulation, where the policy must respond to subtle tool--tissue interactions and visually diverse anatomical environments.
In contrast, open-source surgical videos are relatively abundant and visually diverse. Public datasets and online surgical recordings contain a wide range of surgical scenes, tissue appearances, tool poses, and camera viewpoints. However, these videos generally do not provide synchronized low-level robot action labels, preventing their direct use for robot policy learning.

In this work, we propose \textbf{SurgVIL}, a framework for scaling surgical robot imitation learning using open-source surgical videos. The key idea is to combine two complementary data sources: kinematically labeled robot dataset collected in phantom settings and visually diverse surgical videos. For surgical videos, we generate approximate robot kinematic labels from monocular endoscopic videos as weak supervision in policy training.
SurgVIL is evaluated on two da Vinci surgical robot tasks: needle pick-up and cholecystectomy cutting. For each task, we train policies using either phantom dataset alone or the dataset combining phantom demonstrations with surgical videos. We evaluate three policy backbones, including ACT, $\pi_0$, and Gr00t-H, across multiple visual domains. The results show that adding surgical videos substantially improves generalization beyond the phantom environment, while largely preserving in-domain performance. Our contributions are as follows:

A dual-source surgical imitation learning framework that combines accurately labeled phantom robot demonstrations with visually diverse surgical videos labeled using estimated kinematics; A video-to-action labeling pipeline that converts monocular endoscopic videos into weak robot-action dataset; A task-specific perception module for the needle task that estimates needle position and orientation from endoscopic video frames, enabling an optional input for policy learning; Extensive robot evaluation across two surgical tasks, three policy backbones, and multiple phantom and ex vivo settings.
The dataset and the
action label pipeline are available at \href{https://huggingface.co/datasets/IMERSE-HF/Phantom_dataset_for_cotraining}{huggingface}.

\begin{figure}[t]
    \centering
    \includegraphics[width=1\linewidth]{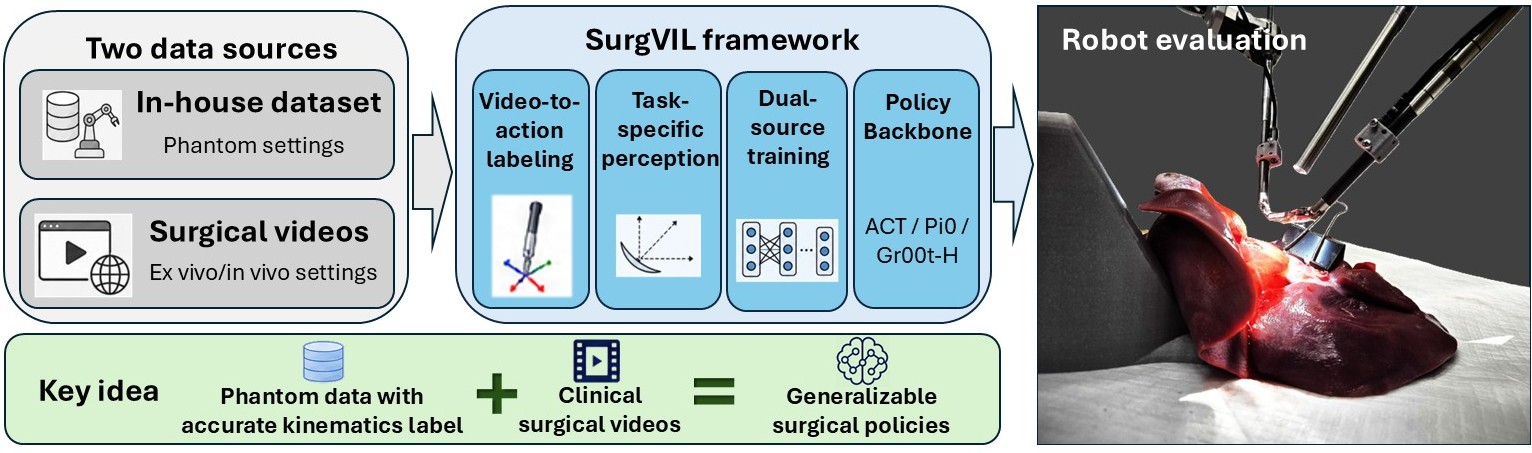}
    \caption{
    Overview of SurgVIL. The proposed framework combines accurately labeled in-house dataset collected in phantom settings with visually diverse open-source surgical videos. The trained policies are evaluated on real-tissue surgical tasks to assess generalization beyond the phantom.
    }
    \label{fig:cover}
    \vspace{-15pt}
\end{figure}

\section{Related Works}

\subsection{Autonomous Robotic Surgery}
Autonomous robotic surgery has been studied as a path toward more consistent and precise procedures. Prior surveys have summarized autonomy levels in surgical robotics from no autonomy to robot assistance and full autonomy~\cite{haidegger2019autonomy,lee2024levels}. Early demonstrations of autonomous soft-tissue surgery showed the feasibility of supervised autonomous suturing, hemostasis, and tissue manipulation~\cite{saeidi2022autonomous, pedram2020autonomous,richter2021autonomous}. These systems demonstrate the potential of surgical autonomy, but often rely on specialized sensing, planning, or control.
More recently, learning-based approaches have been explored for surgical robot autonomy. Surgical Robot Transformer (SRT) and SRT-H showed that imitation learning can be applied to da Vinci surgical manipulation tasks by using visual observations and relative robot actions~\cite{kim2025surgical, kim2025srt}. In parallel, reinforcement-learning and simulation-based methods have investigated demonstration-guided exploration, decision-transformer policies, and generalized sim-to-real task autonomy for surgical robots~\cite{huang2023guided,fu2024multi,long2025surgical}. These works show the promise of learning-based surgical autonomy, but scalable data collection remains a central bottleneck.
Recent work has begun to explore surgical robot policy learning from simulated videos using world models~\cite{he2025surgworld}. While this direction is promising, such approaches require substantial computational resources to realistically model clinical settings. Policies trained in simulation also suffer from a sim-to-real gap, and robust transfer to real surgical robot systems remains difficult. In contrast, our work focuses on using existing open-source surgical videos to improve surgical robot policy generalization.


\subsection{Robot Imitation Learning Policies}
Imitation learning has become a strong paradigm for robot manipulation, especially when expert demonstrations are available. Action Chunking Transformer (ACT) predicts future action chunks and temporally ensembles them, improving consistency for fine-grained manipulation~\cite{zhao2023learning}. Large-scale imitation-learning policies show that transformer and diffusion backbones can benefit from pretraining on diverse robot trajectories~\cite{liu2025rdt}. 
Vision-language-action (VLA) models have recently emerged as general-purpose robot policy backbones. RT-2 showed that vision-language pretraining can be transferred to robot control by representing actions as tokens~\cite{zitkovich2023rt}. $\pi_0$ uses a pretrained vision-language model backbone and a flow-matching action expert to produce continuous robot actions~\cite{black2024pi_0}. OpenVLA demonstrates open-source VLA training on diverse robot demonstrations~\cite{kim2024openvla}. In the medical robotics domain, GR00T-H is a medical robotics foundation model trained on Open-H-Embodiment~\cite{nelson2026open}. These models show that large-scale pretraining can improve robot policy learning, but their performance still depends on the relevance of downstream tasks and domains. Rather than proposing a new policy architecture, we study how to improve surgical robot policy generalization based on these pretrained models.

\subsection{Learning Robot Policies from Videos}
A growing body of work in general robotics has explored how large-scale human or internet videos can improve robot learning. Many works, including R3M ~\cite{nair2023r3m} and VIP~\cite{ma2022vip}, use videos primarily for general representation or reward pretraining. 
Other works use human videos more directly for policy learning, either by extracting plans from human videos, pretraining generative robot policies on videos, or co-training policies with human and robot demonstrations~\cite{wu2024unleashing,kareer2025egomimic}. More recent methods further combine trajectories derived from human videos with robot data to reduce the cost of data collection~\cite{liu2025immimic}. These works show that visually diverse videos can provide useful information for robot policy learning. 
However in this study, the domain gap is not between human body and robot, but between controlled phantom settings and realistic surgical scenes. We therefore estimate tool kinematics from surgical videos and combine these labeled videos with phantom robot dataset for policy training.


\section{Methods}
\subsection{Dataset Construction}

Two representative surgical tasks are considered in this study: needle pick-up and cholecystectomy cutting. These tasks are selected because they involve visually guided object interaction, precise control, and distinct surgical scenarios. Both tasks are executed using a da Vinci surgical robot.
In the needle pick-up task (needle task), a large needle driver localizes, grasps, and lifts a surgical needle from the surface. In the cholecystectomy cutting task (chole task), curved scissors approach the clipped target tissue, align with the cutting site, and execute a cutting motion while avoiding other tissue. Three surgical clips have been placed on the target tubular structure, which may be either the cystic duct or the cystic artery. For each task, two types of datasets are constructed: an ``in-house dataset'' and a ``video dataset''.




The in-house datasets were collected using a dVRK-Si robot under controlled experimental conditions, as shown in Figure~\ref{fig:cover}. For the needle task, 1,000 demonstrations were performed on a suture pad using a large needle driver. For the chole task, 700 demonstrations were collected on a gallbladder phantom using curved scissors. During data collection, the robot was teleoperated to complete each task while synchronized endoscopic images, wrist-camera images, and robot kinematic data were recorded. Because the kinematic labels were obtained directly from the robot system, they were treated as accurate action labels. To increase dataset diversity, the poses of the phantom, needle, lighting conditions, and the initial pose of the surgical tool were frequently adjusted during data collection. However, because these datasets were collected on phantoms, their visual appearance differs from the ex vivo and in vivo surgical scenes.

Open-source surgical videos contain the endoscopic view but do not initially provide synchronized robot kinematics. Therefore, we generate approximate kinematic labels from video using the proposed labeling pipeline described in Section~\ref{sec:video-to-action} to build the video dataset. Although the video-derived labels are less accurate than labels directly from robot, they provide more informative supervision than using videos alone and can serve as useful augmentation during policy training.
For needle pick-up, the video dataset was primarily curated from the SurgVU dataset~\cite{zia2025surgical}, SAR-RARP50~\cite{psychogyios2023sar} showing da Vinci needle pick-up under endoscopic view, resulting in 500 episodes.
For cholecystectomy cutting, the video dataset was curated from videos in the ImitateCholec dataset~\cite{kim2025srt}, resulting in 700 episodes. The video dataset includes diverse clinical settings and provides more realistic tissue appearance and tool--tissue interactions than the in-house phantom dataset.

\subsection{Tool and Needle Segmentation}
For policies evaluated on the needle task, we additionally consider an optional needle pose input. Policies with the ``\_PO'' suffix receive the 2D needle center and the local tangent direction at that point. These are estimated from the first frame of each demonstration and are kept fixed throughout the episode. To obtain this information, we train a lightweight network based on U-net~\cite{ronneberger2015u} to segment the surgical tool and needle. Details of the segmentation work are provided in the Appendix.
The tool segmentation mask is used in the video-to-action labeling pipeline, whereas the needle segmentation mask is further processed using a curve-fitting algorithm to remove the attached suture thread and estimate the needle center point and the tangent direction.

\begin{figure}[tb]
    \centering
    \includegraphics[width=0.95\linewidth]{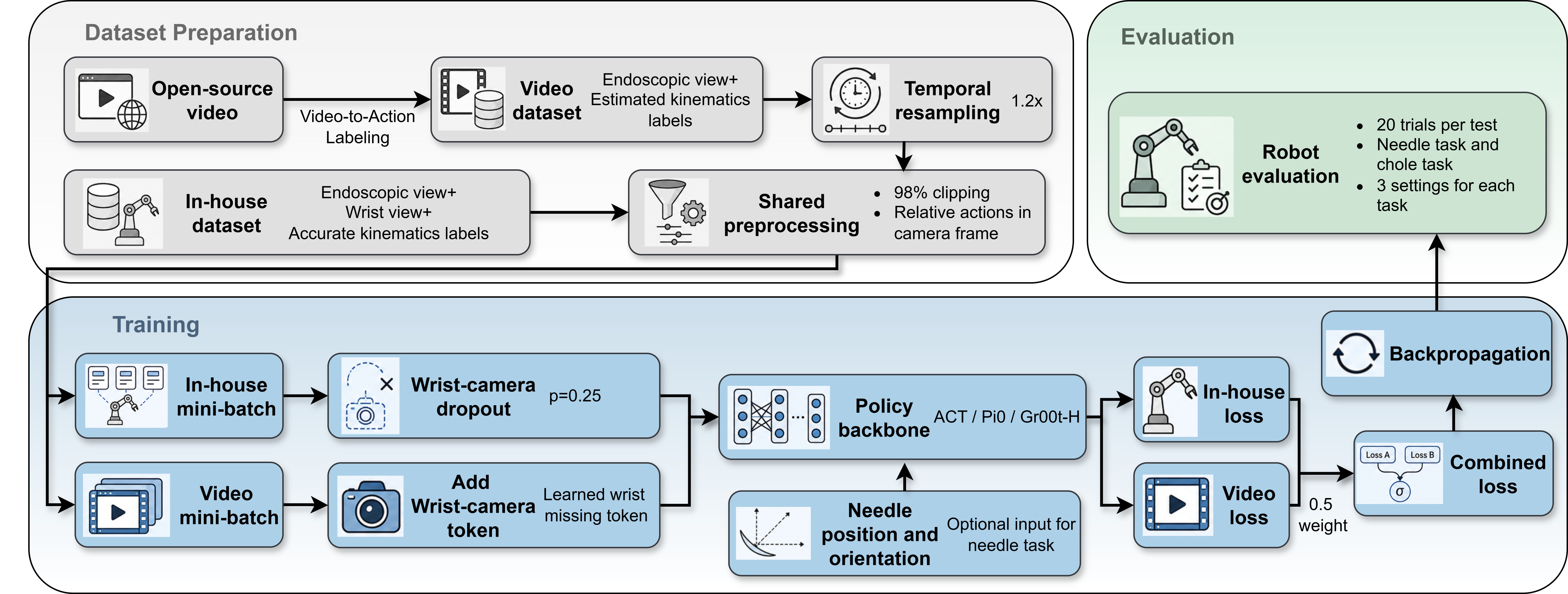}
    \caption{
    Flow chart for the dataset preparation, dual-source policy training and evaluation
    }
    \label{fig:flow_chart}
    \vspace{-15pt}
\end{figure}

\subsection{Video-to-Action Labeling Pipeline}
\label{sec:video-to-action}

The video-to-action labeling pipeline is built on a differentiable rendering-based framework for estimating surgical tool kinematics from monocular surgical videos. Given a monocular video, the pipeline uses a trained differentiable rendering policy to optimize the pose of a renderable tool model such that the rendered tool appearance matches the input frame.
Although prior differentiable rendering methods, such as SurgiPose~\cite{chen2025surgipose}, provide an effective way to extract kinematics from surgical videos, they may produce position errors greater than 1 cm, temporally inconsistent pose estimates, and do not support all required surgical instruments. We therefore extend its applicability to additional surgical instruments and introduce a new design that improves the accuracy. 
\vspace{-0.2em}
\begin{equation}
\label{eq:loss}
\mathcal{L}_{t}
=
\lambda_{\mathrm{ssim}}
\left(
1-\operatorname{SSIM}
\left(
I_t^{\mathrm{ren}}, I_t^{\mathrm{obs}}
\right)
\right)
+
\lambda_{\mathrm{dice}}
\mathcal{L}_{\mathrm{Dice}}
\left(
M_t^{\mathrm{ren}}, M_t^{\mathrm{obs}}
\right)
+
\lambda_{\mathrm{temp}}
\rho
\left(
\xi_t-\xi_{t-1}
\right),
\end{equation}
Eq.~\ref{eq:loss} shows the loss function with a temporal term. Here $I_t^{ren}$ and $I_t^{obs}$ denote the rendered and observed images. $M_t^{ren}$ and $M_t^{obs}$ are the segmentation masks. The estimated tool pose is represented as $T_t \in SE(3)$. 
Here, SSIM loss measures the structural similarity, and the Dice loss compares the segmentation masks. The vector $\xi_t=log(T^{-1}_{t-1}T_t)$ represents the relative rigid-body motion between two frames in the Lie algebra. The temporal term penalizes abrupt changes in acceleration, encouraging smooth trajectories over time. We use the Charbonnier penalty $
\rho(x)=\sqrt{\|x\|_2^2+\epsilon^2}$. 
We further correct the systematic depth bias using a post-hoc calibration model, which fits an affine mapping between the estimated depth $z_{est}$ and the ground-truth depth $z_{gt}$: $
z_{gt} \approx a z_{est} + b$.
The calibration parameters $a$ and $b$ are estimated once using synthetically generated MuJoCo training data and are subsequently kept fixed. Evaluation on the independent SurgRIPE dataset therefore measures transfer of the complete pose-estimation pipeline, including depth calibration, to unseen real surgical images.

The final calibrated pose sequence is converted into action labels by computing frame-to-frame relative motion. These video-derived action labels are used as weak supervision for policy training. We also extended this pipeline to support the curved scissors instrument using a custom CAD model, which is required for the chole task.

\subsection{Dual-source Policy Training}\label{subsec:dual}

We train the policies using two datasets described above: the in-house dataset and the video dataset, as shown in Figure~\ref{fig:flow_chart}. This design encourages the policy to learn task execution from the in-house dataset while learning tool and tissue appearance variations from the video dataset.

Before training, we apply a shared action normalization to both datasets. To reduce the effect of outliers, particularly in the video dataset, each action component is clipped at the 98th percentile before normalization. The video dataset is temporally resampled with a speed factor of 1.2 to better match the motion speed distribution of the in-house dataset.

The tasks are formulated as a conditional action prediction problem. At time step $t$, the policy receives a center endoscopic image $I^{e}_{t}$, a wrist camera image $I^{w}_{t}$, and a task label input $l_{t}$. To unify the input format across datasets, we replace the unavailable wrist image in the video dataset with a learned missing-view token $m^{w}$. During training on the in-house dataset, we additionally apply wrist-camera dropout with probability $0.25$, replacing with the same token $m^{w}$. This encourages robustness to missing views and reduces over-reliance on the wrist camera. The policy $f_{\theta}$ predicts the relative robot action $\hat{\mathbf{a}}_{t}$ as
\begin{equation}
    \hat{\mathbf{a}}_{t}
    =
    f_{\theta}
    \left(
        I^{e}_{t},
        \tilde{I}^{w}_{t},
        l_{t}
    \right), \quad \quad
   \hat{\mathbf{a}}_{t}
    =
    \left[
        \Delta \mathbf{p}_{t},
        \Delta \mathbf{q}_{t},
        g_{t}
    \right]
    \label{eq:policy_function}
\end{equation}
where $\tilde{I}^{w}_{t}$ is either $I^{w}_{t}$ or $m^{w}$, $\Delta \mathbf{p}_{t} \in \mathbb{R}^{3}$ is the translational displacement, $\Delta \mathbf{q}_{t} \in \mathbb{R}^{4}$ is the rotation quaternion, and $g_{t}$ is the jaw angle. The input order is fixed, with the endoscopic image first, followed by the wrist-camera input.

At each training step, we independently sample two mini-batches of equal size from both datasets. The two mini-batches are processed by the same policy network with shared parameters. Let $\mathcal{B}_{h}$ denote the in-house mini-batch and $\mathcal{B}_{v}$ denote the video mini-batch. The training objective is a weighted combination of the two losses:
\begin{equation}
    \mathcal{L}
    =
    \mathcal{L}_{policy}(\mathcal{B}_{h})
    +
    \lambda_{v}\mathcal{L}_{policy}(\mathcal{B}_{v}),
    \label{eq:dual_source_loss}
\end{equation}
where $\mathcal{L}_{policy}$ is the imitation loss used by each backbone, and $\lambda_{v}$ controls the contribution of video dataset. We set $\lambda_{v}=0.5$, selected from candidate values $\{0.1, 0.25, 0.5, 0.75, 1.0\}$ based on validation loss. This weighting keeps the in-house dataset as the primary source of accurate action supervision while allowing the policy to benefit from the visual diversity of the video dataset.

We use the same dual-source training procedure for all evaluated policy backbones: ACT~\cite{zhao2023learning}, $\pi_0$~\cite{black2024pi_0}, and Gr00t-H~\cite{nelson2026open}. These models are used as representative imitation-learning and vision-language-action policy architectures. ACT is used as standard imitation learning backbone; $\pi_0$ provides a general-purpose VLA policy backbone; while Gr00t-H provides a medical foundation VLA backbone.
For all three backbones, the policy input consists of the endoscopic image, the wrist-camera input or missing-view token, and the task label. The output action follows the same relative action format defined in Eq.~\ref{eq:policy_function}. All implementation details specific to each backbone are provided in the Appendix.

\section{Experiments and Results}

\subsection{Experiment Setup}
\begin{figure}[b]
    \centering
    \includegraphics[width=0.65\linewidth]{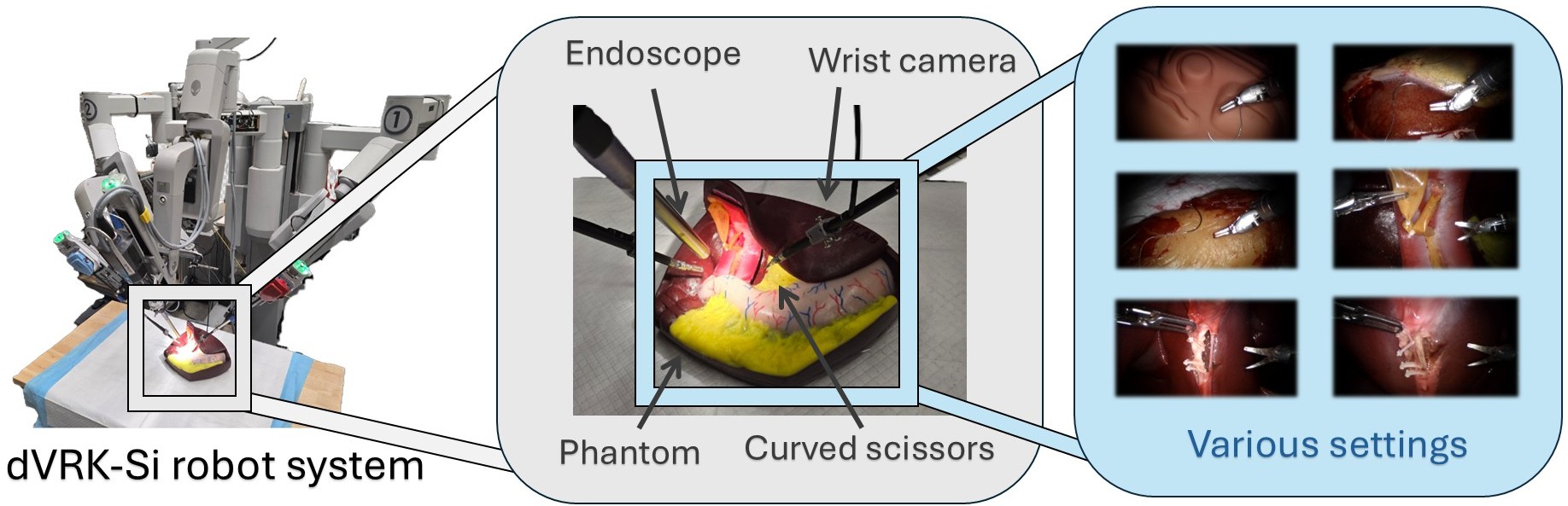}
    \caption{
    Experimental Setup. The evaluation was conducted on a dVRK-Si robot system, with an additional wrist camera mounted on the surgical instrument.
    }
    \label{fig:setup}
    \vspace{-15pt}
\end{figure}

The experiments are designed to answer four questions. First, does adding surgical video data improve policy generalization to visually realistic and out-of-domain surgical scenes? Second, is the improvement due to the visual diversity of the video data rather than only the increased number of training episodes? Third, do the estimated action labels provide useful supervision beyond exposure to the visual diversity of the surgical videos alone? Fourth, is the benefit consistent across different policy backbones and tasks?

All policies are evaluated on a dVRK-Si surgical robot, as shown in Figure~\ref{fig:setup}.
For the needle task, the robot uses a large needle driver to grasp and lift a surgical needle. Policies are evaluated on three settings: suture pad, porcine liver and gallbladder, and chicken breast with fake blood. The suture pad matches the in-house dataset environment. The porcine liver and gallbladder setting is visually similar to the video dataset, while the chicken breast with fake blood setting represents a more challenging out-of-domain background.
For the chole task, the robot uses curved scissors to cut the target tissue region. Similarly, policies are evaluated in three settings: gallbladder phantom, normal ex vivo gallbladder, and abnormal ex vivo gallbladder. The phantom setting matches the in-house training environment. The normal ex vivo setting is considered in-domain with respect to the video dataset, as the cystic duct appears on the left side of the camera view in both settings. The abnormal ex vivo setting includes cases where the cystic duct and cystic artery are either spatially switched or not fully dissected, evaluating out-of-domain generalization.

\subsection{Video-to-Action Labeling Accuracy}

We evaluated the video-to-action labeling pipeline on robot datasets with kinematic labels to assess the accuracy and generalization of the pose estimation results. The large needle driver model was evaluated on the SurgRIPE dataset~\cite{xu2025surgripe}, while the curved scissors model was evaluated on the ImitateCholec dataset~\cite{kim2025srt}. The large needle driver achieves $6.3 \pm 1.7$ mm position error and $4.12 \pm 1.55^\circ$ orientation error. This result is overall better than the methods in the SurgRIPE challenge. For the curved scissors, the error is $9.5 \pm 2.8$ mm and $7.01 \pm 2.43^\circ$. The error comparison with earlier methods are in the appendix and additional visualizations are provided in Appendix.

\subsection{Training Configurations}
We evaluate three policy backbones: ACT, $\pi_0$, and Gr00t-H. For each backbone, we train policies using either only the in-house dataset or a combination of the in-house dataset and the video dataset. The notation $(x+y)$ denotes training with x episodes of in-house dataset and y episodes of video dataset.
For the needle task, we evaluate $(500+0)$ and $(500+500)$ for all three backbones. To isolate the contribution of the estimated action labels from the visual diversity of the video dataset, we additionally train a matched zero action-label ablation, denoted $\_NA$. These models use the same 500 in-house and 500 surgical-video episodes as the \((500+500)\) setting, but all action labels associated with the video episodes are set to zero. To separate the effect of visual diversity from the effect of simply increasing the number of training episodes, we additionally train a $(1000+0)$ setting, using the same total number of episodes as $(500+500)$. We also evaluate an auxiliary input, denoted as $\_PO$, where the policy receives the needle center point and its tangent vector estimated from the first frame. This input is not updated during execution.
For the chole task, we evaluate $(700+0)$ and $(700+700)$ for all three backbones. All policies are trained using the same action representation, observation format, and dual-source training procedure described in Sec.~\ref{subsec:dual}.

Each policy is evaluated for 20 independent trials in each test setting. The primary metric is the number of successful trials out of 20, and corresponding success rate. For the needle task, a trial is considered successful if the robot grasps the needle and lifts it from the surface. For the chole task, a trial is considered successful if the robot completes the intended cutting motion on the target tissue without hurting non-target tissue. If a trial does not complete the task within 40 seconds or exhibits behavior likely to damage the tissue, it is manually terminated and counted as a failure. During evaluation, all policies are tested sequentially under the same settings. The tissue pose and robot starting pose are then changed for the next evaluation round to ensure a fair comparison across policies.

\begin{figure}[tb]
    \centering
    \includegraphics[width=0.75\linewidth]{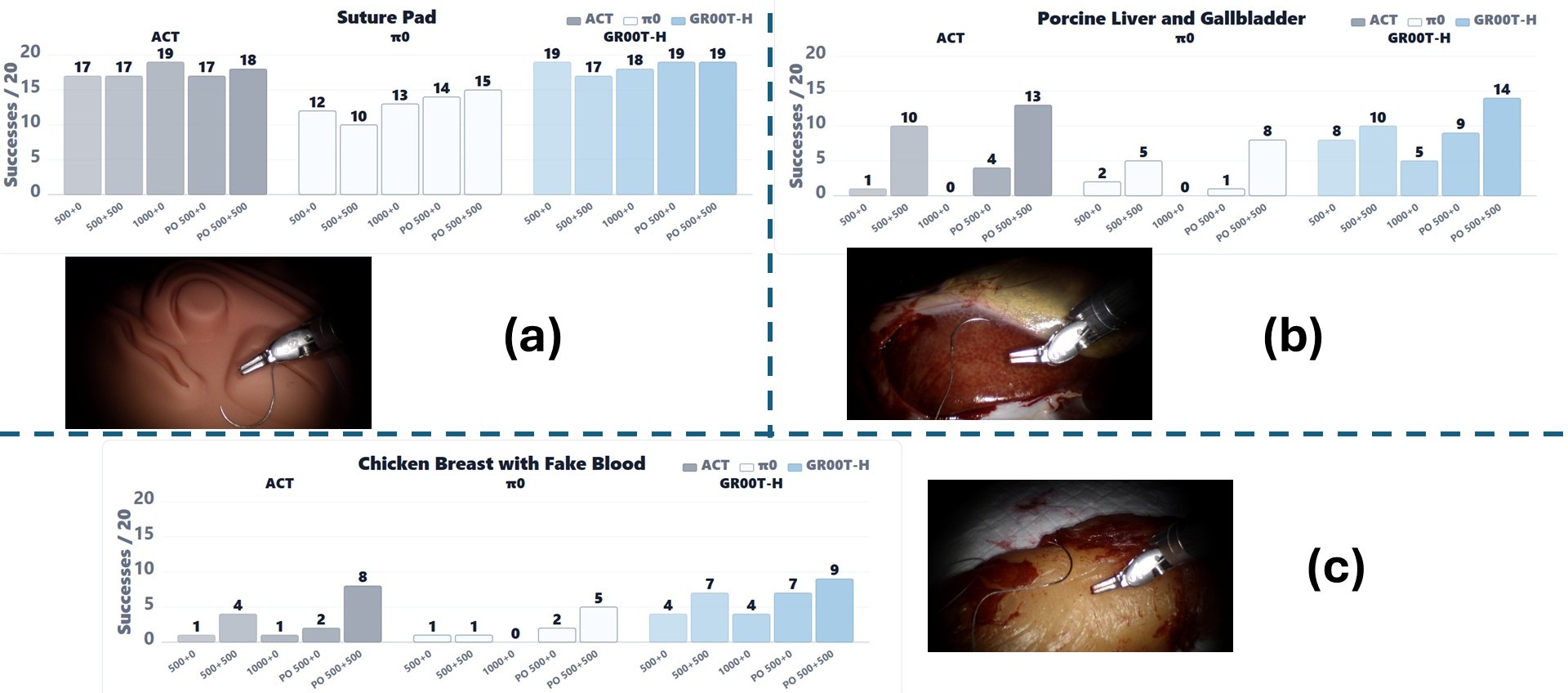}
    \caption{
    Needle task results on (a) Suture pad, (b) Porcine liver and gallbladder, and (c) Chicken breast with fake blood. The results are also listed as table in the Appendix.
    }
    \label{fig:needle_results}
    \vspace{-15pt}
\end{figure}

\subsection{Evaluation on Needle Task}

Figure~\ref{fig:needle_results} reports the needle task results. Comparing the six matched policy pairs trained with $(500+0)$ and $(500+500)$, adding the video dataset substantially improves average generalization to ex vivo and out-of-domain settings. The average success rate increases by 29.2\% on porcine liver and gallbladder, and by 14.2\% on chicken breast with fake blood. The average success rate on the suture pad decreases by only 1.7\%, indicating that adding video data improves generalization while largely preserving performance in the phantom environment.
The optional needle state input further improves performance. Across the six matched comparisons, adding the initial needle state increases the average success rate by 8.3\% on the suture pad, 10.8\% on porcine liver and gallbladder, and 12.5\% on chicken breast with fake blood. These results show that the initial needle position and orientation provide useful task information. However, this input does not replace the benefit of visually diverse video training data. Under the same input setting, policies trained with $(500+500)$ still outperform their $(500+0)$ counterparts on tissue-based backgrounds. 

The zero action-label ablation further shows that visual exposure alone does not explain the improvement from surgical video training. Averaged across the two ex vivo settings and three backbones, the NA models improve performance by only 2.5 percentage points over the $(500+0)$ models. In contrast, the original $(500+500)$ models trained with estimated action labels achieve 16.7 percentage point improvement. These results indicate that the estimated kinematic labels provide useful supervision beyond visual diversity alone.

Importantly, increasing the amount of in-house phantom data alone does not produce the same improvement. Comparing $(500+0)$ and $(1000+0)$ across all backbones, adding more in-house data slightly improves performance on the suture pad but decreases performance on other settings. This suggests that additional phantom data may further specialize the policy to the phantom environment rather than improve visual generalization. Therefore, the improvement cannot be explained by the dataset size alone.

\subsection{Evaluation on Chole Task}

\begin{figure}[tb]
    \centering
    \includegraphics[width=0.7\linewidth]{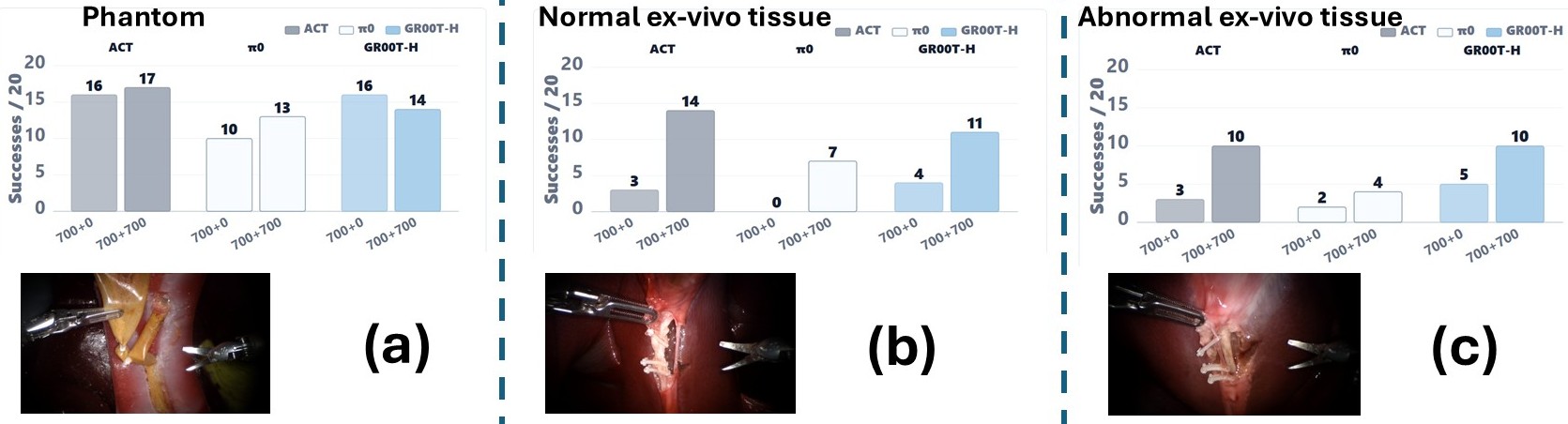}
    \caption{
    Chole task results on (a) Phantom, (b) Normal ex vivo tissue, and (c) abnormal ex vivo tissue. The results are also listed as table in the Appendix.
    }
    \label{fig:chole_results}
    \vspace{-15pt}
\end{figure}

Figure~\ref{fig:chole_results} reports the chole task results. Comparing the three policies trained with $(700+0)$ and the three policies trained with $(700+700)$, adding the video dataset consistently improves performance on both ex vivo evaluation settings. The average success rate increases by 41.7\% on normal ex vivo tissue and by 23.3\% on OOD tissue. The average success rate on the gallbladder phantom increases slightly by 3.3\%. This indicates that adding the video dataset substantially improves generalization to realistic tissue settings while largely preserving performance in the phantom environment.
The improvement is consistent across all three policy backbones, showing that the benefit of video dataset is not limited to a specific policy architecture. Moreover, because the chole task uses a different instrument and requires direct interaction with deformable tissue, the results suggest that video dataset can improve policy generalization beyond a single task. We did not include the $(1400+0)$ setting in the chole task, as the needle task's results showed that increasing the amount of in-house data alone did not improve performance.

\section{Discussion and Limitations}
Overall, the results support the central hypothesis of SurgVIL, that surgical videos can serve as a scalable source for surgical robot policy learning. Across two surgical tasks, three policy backbones, and multiple visual domains, adding surgical videos substantially improves generalization to real-tissue and out-of-distribution settings. The results also show that this gain cannot be explained by data volume or visual diversity alone. Considering the three policy backbones, ACT benefits most from video data. This suggests that video can substantially improve a standard imitation learning backbone. Gr00t-H shows smaller relative gains likely due to prior surgical pretraining on large surgical robot dataset. But adding the video dataset still improves its generalization in most ex vivo settings. $\pi_0$ achieves relatively lower performance than other policies. These results suggest that general robot pretraining alone may be insufficient for precise surgical manipulation tasks and that larger surgical training dataset is required for large model to bridge the domain gap.

Qualitative inspection of the ex vivo trials suggests that policies only trained with phantom dataset frequently fail because of target localization and depth estimation errors. In the needle task, the tool may approach an incorrect location or depth, while in the chole task it may approach the wrong tubular structure. These behaviors appeared less frequently after adding surgical videos.

This work has several limitations. First, although the proposed labeling pipeline achieves better performance than prior methods, the kinematic labels still contain noise, which can limit policy performance.
Second, this work focuses on demonstrating the value of combining surgical video data. Although the results show clear gains, the final success rates in ex vivo and OOD settings are still not close to saturation. Further improvements in label accuracy, policy architecture, and larger-scale surgical video curation are still needed to achieve higher success rates.
Third, the two evaluated tasks are relatively short-horizon surgical tasks. Needle pick-up and cholecystectomy cutting were selected because they require precise visually guided manipulation and allow evaluation across different instruments, tissues, and target configurations, but the current results do not establish performance on long-horizon surgical procedures. Extending SurgVIL to longer procedures may require adding a high-level policy or learning from longer video sequences.
Fourth, the experiments are conducted on the same dVRK-Si system. It does not test transfer across different robot systems, camera, or operating room setups.
Finally, the evaluation uses 20 trials per condition. Although this provides a practical measure of the real robot performance, evaluations with more trials would provide stronger quantitative analysis.


\section{Conclusion}

In this work, we presented SurgVIL, a dual-source imitation learning framework for surgical robot policy learning. SurgVIL combines kinematically labeled phantom robot data with visually diverse open-source surgical videos for policy learning. This design addresses a central data bottleneck in surgical robot learning: datasets with robot action are difficult to collect in realistic surgical settings, while open-source surgical videos are relatively abundant but lack kinematics. We evaluated SurgVIL on two da Vinci surgical robot tasks: needle pick-up and cholecystectomy cutting. Across ACT, $\pi_0$, and Gr00t-H backbones, adding surgical video improved policy generalization to ex vivo and out-of-domain settings while largely preserving performance in the phantom environment. The needle task further showed that the improvement cannot be explained by additional data volume alone. These results support that open-source surgical videos can serve as a scalable source of useful training data for surgical robot policy learning.


\clearpage
\acknowledgments{This research has been generously funded by the ARPA-H 75N91023C00048.}


\bibliography{example}  

\section*{Appendix}

\subsection*{Policy Training Details}

\textbf{ACT:} 

Action Chunking Transformer (ACT) is used as a non-VLA imitation learning baseline. ACT predicts a sequence of future actions from visual observations, which allows the policy to model temporally correlated robot motions. Since ACT does not rely on a pretrained  model, it provides a useful reference for evaluating whether the proposed dataset construction benefits standard visuomotor imitation learning.

\begin{table}[H]
\centering
\caption*{ACT hyperparameters}
\begin{tabular}{lll}
\hline
\textbf{Parameter} & \textbf{ACT for needle task} & \textbf{ACT for chole task} \\
\hline
Model / initialization & ACT & ACT \\
Optimizer & AdamW & AdamW \\
Learning rate & $5 \times 10^{-4}$ & $5 \times 10^{-4}$ \\
Batch size & 256 & 256 \\
Adam $\beta_1$ & 0.9 & 0.9 \\
Adam $\beta_2$ & 0.999 & 0.999 \\
Adam $\epsilon$ & $1 \times 10^{-8}$ & $1 \times 10^{-8}$ \\
Weight decay & $1 \times 10^{-4}$ & $1 \times 10^{-4}$ \\
Learning-rate schedule & LambdaLR & LambdaLR \\
Warmup & 750 steps & 1,000 steps \\
Training steps & 15,000 & 20,000 \\
Hidden dimension & 512 & 512 \\
Feedforward dimension & 3200 & 3200 \\
Training chunk size & 60 & 60 \\
Evaluation action horizon & 20 & 20 \\
KL weight / $\beta$ & 10 & 10 \\
Data FPS & 30 Hz & 30 Hz \\
\hline
\end{tabular}
\end{table}

\textbf{$\pi_0$:}

$\pi_0$ is a vision-language-action model designed for general robot control. It builds on a pretrained vision-language model backbone and adds an action expert that predicts continuous action chunks using flow matching. This model is included to evaluate whether a general-purpose VLA policy can benefit from surgical videos.

\begin{table}[H]
\centering
\caption*{$\pi_0$ fine-tuning hyperparameters}
\begin{tabular}{lll}
\hline
\textbf{Parameter} & \textbf{$\pi_0$ for needle task} & \textbf{$\pi_0$ for chole task} \\
\hline
Model / initialization & $\pi_0$ & $\pi_0$ \\
Optimizer & AdamW & AdamW \\
Learning rate & $1 \times 10^{-4}$ & $1 \times 10^{-4}$ \\
Batch size & 128 & 128 \\
Adam $\beta_1$ & 0.9 & 0.9 \\
Adam $\beta_2$ & 0.95 & 0.95 \\
Adam $\epsilon$ & $1 \times 10^{-8}$ & $1 \times 10^{-8}$ \\
Weight decay & 0.1 & 0.1 \\
Learning-rate schedule & Cosine & Cosine \\
Warmup & 1,000 steps & 1,500 steps \\
Training steps & 11,000 & 15,000 \\
Full fine-tuning & True & True \\
Training chunk size & 50 & 50 \\
Evaluation action horizon & 20 & 20 \\
Data FPS & 30 Hz & 30 Hz \\
\hline
\end{tabular}
\end{table}

\textbf{Gr00t-H:}

Gr00t-H denotes Gr00t N1.6 further fine-tuned on a large surgical robot dataset. Similar to $\pi_0$, it serves as a pretrained VLA-style policy backbone. Comparing Gr00t-H with ACT and $\pi_0$ allows us to test whether the benefit of adding surgical videos is consistent across both non-VLA and pretrained VLA policy families.

\begin{table}[H]
\centering
\caption*{GR00T-H fine-tuning hyperparameters}
\small
\begin{tabular}{lll}
\hline
\textbf{Parameter} &
\textbf{GR00T-H for needle task} &
\textbf{GR00T-H for chole task} \\
\hline
Model / initialization & GR00T-H & GR00T-H \\
Optimizer & AdamW & AdamW \\
Learning rate & $2 \times 10^{-5}$ & $2 \times 10^{-5}$ \\
Batch size & 64 & 64 \\
Weight decay & $1 \times 10^{-5}$ & $1 \times 10^{-5}$ \\
Learning-rate schedule & Cosine decay & Cosine decay \\
Warmup & 5\% & 5\% \\
Training steps & 25,000 & 35,000 \\
Backbone tuning & LLM backbone frozen & LLM backbone frozen \\
Action horizon & 50 & 50 \\
Action normalization & Mean/std & Mean/std \\
Random crop & 95\% & 95\% \\
Rotation augmentation & $\pm 5^\circ$ & $\pm 5^\circ$ \\
Coarse pixel dropout & 0.5 & 0.5 \\
Color jitter brightness & 0.12 & 0.12 \\
Color jitter contrast & 0.15 & 0.15 \\
Color jitter saturation & 0.15 & 0.15 \\
Color jitter hue & 0.02 & 0.02 \\
Multi-view dropout & 0.25 & 0.25 \\
Gradient clipping & 1.0 & 1.0 \\
\hline
\end{tabular}
\end{table}

\subsection*{Video-to-Action Labeling Details and Results}

\begin{table}[H]
\centering
\caption*{Hyperparameters used in the video-to-action labeling loss.}
\begin{tabular}{lcc}
\hline
Parameter & Description & Value \\
\hline
$\lambda_{\mathrm{ssim}}$ & Weight for SSIM image similarity loss & $0.5$ \\
$\lambda_{\mathrm{dice}}$ & Weight for Dice mask loss & $1.0$ \\
$\lambda_{\mathrm{temp}}$ & Weight for temporal consistency loss & $0.05$ \\
$\epsilon$ & Numerical stability constant in $\rho(\cdot)$ & $1 \times 10^{-6}$ \\
\hline
\end{tabular}
\end{table}


The large needle driver model was evaluated on the SurgRIPE dataset~\cite{xu2025surgripe}. We chose this dataset because it also includes a pose estimation challenge, enabling direct comparison with state-of-the-art results. The curved scissors model was evaluated on the ImitateCholec dataset~\cite{kim2025srt}. Since all training data of the proposed method are synthetically generated in MuJoCo and the coefficients in post-hoc depth calibration is also calculated on this dataset, evaluating the model on real surgical videos does not introduce test set leakage.

\begin{table}[H]
\centering
\caption*{Comparison of pose estimation accuracy on large needle driver}
\begin{tabular}{lcc}
\hline
Method & Position Error (mm) & Orientation Error ($^\circ$) \\
\hline
SurgiPose~\cite{chen2025surgipose} & $12.0$ & Not reported \\
TUDU~\cite{xu2025surgripe} & $5.8$ & $21.48$ \\
IGTUM~\cite{xu2025surgripe} & $\textbf{5.4}$ & $10.71$ \\
UOL~\cite{xu2025surgripe} & $8.3$ & $17.73$ \\
FoundPose~\cite{ornek2024foundpose} & $20.7$ & $55.01$ \\
MegaPose~\cite{labbe2022megapose} & $13.4$ & $44.09$ \\
Ours method & $6.3$ & $\textbf{4.12}$ \\
\hline
\end{tabular}
\end{table}

This table reports the pose estimation error for the large needle driver of our method, SurgiPose, top three models from SurgRIPE , and two zero-shot methods. Our method achieves $6.3 \pm 1.7$ mm position error, which is comparable to the IGTUM. Our method achieves an orientation error of $4.12 \pm 1.55^\circ$, which is substantially lower than IGTUM's $10.71^\circ$ and other methods. Accurate orientation estimation is important for both the needle task and chole task. Moreover, among the compared methods, only our method and SurgiPose are specifically designed for pose estimation in videos, whereas the other methods are developed for single-image. As a result, these methods are much slower when applied to videos and tend to produce temporally inconsistent pose estimates across frames. Considering both overall accuracy, computational cost and temporal consistency, our proposed method achieves the best overall performance in this setting.

For the curved scissors, the error is $9.5 \pm 2.8$ mm and $7.01 \pm 2.43^\circ$. The larger error observed for the curved scissors may be due to the use of a manually measured and custom-built CAD model, whereas an accurate CAD model was available for the large needle driver. To the best of our knowledge, we didn't find prior work on kinematic estimation for curved scissors. We provide an overlay visualization on the original frame as shown in the figure.

\begin{figure}[H]
    \centering
    \includegraphics[width=1\linewidth]{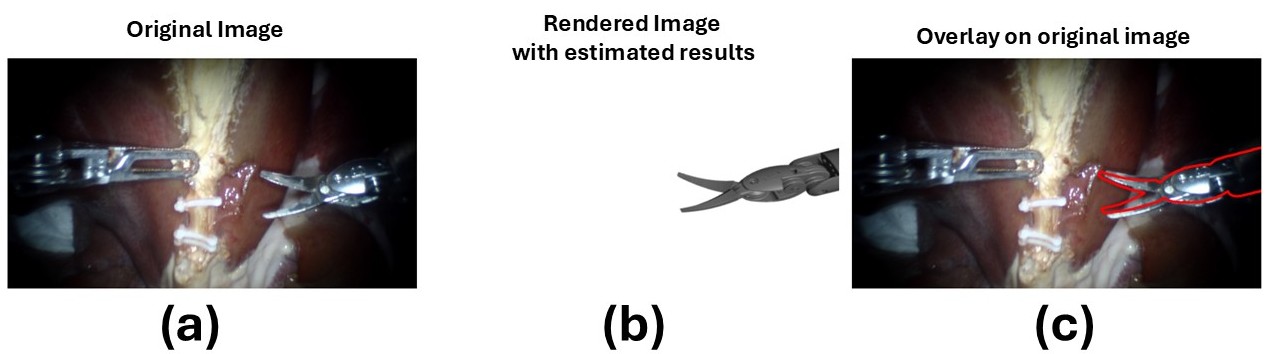}
    \caption*{
    Visualization of kinematic estimation results for curved scissors. (a) Original image, (b) Rendered image with estimated kinematics results from Mujoco, (c) Overlay the rendered outline on original image.
    }
\end{figure}

\subsection*{Tool and Needle Segmentation Network}

To extract the image evidence required by our downstream pipeline, we train a lightweight U-Net variant to segment both the surgical tool and the needle from input endoscopic image. While the overall architecture retains the encoder plus decoder structure and skip connections of the standard U-Net~\cite{ronneberger2015u}, we modify the encoder to improve computational efficiency. The original U-Net encoder is replaced with a MobileNetV2 backbone~\cite{sandler2018mobilenetv2}, whose inverted residual blocks and depthwise separable convolutions substantially reduce the number of convolutional operations compared with standard convolution layers. This design is motivated by prior surgical needle localization work~\cite{pryor2021localization}.

The network is trained using 60 surgical videos annotated with SAM2~\cite{ravi2025sam}. The annotations provide pixel-wise labels for the surgical tool and needle in each frame, which are used as supervision for the segmentation network. Because the needle is much smaller than both the background and the tool, the training data has strong class imbalance. To address this, we optimize the network using a weighted combination of multi-class cross-entropy loss and foreground Dice loss. The cross-entropy term encourages correct pixel-wise classification across all classes, while the Dice term directly improves overlap between the predicted and ground-truth foreground regions. Applying the Dice loss to the tool and needle classes reduces the tendency of ignoring small foreground structures.

The tool segmentation mask is used in the video-to-action labeling pipeline, whereas the needle segmentation mask is further processed using a curve-fitting algorithm to remove the attached suture thread and estimate the needle center point and the tangent direction, as shown in the figure.

\begin{figure}[H]
    \centering
    \includegraphics[width=0.5\linewidth]{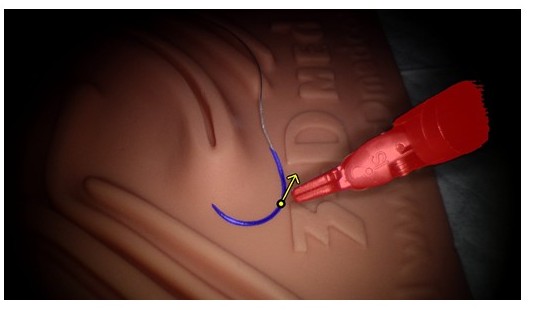}
    \caption*{
    Segmentation masks of the surgical tool and needle, together with estimated needle center point and tangent vector.
    }
\end{figure}

\subsection*{Full Experimental Results}
\begin{table}[H]
\centering
\caption*{Full needle task results. Each entry reports successful trials out of 20, with success rate in parentheses.}
\resizebox{\textwidth}{!}{
\begin{tabular}{lcccc}
\hline
Backbone & Training setting & Suture pad & Porcine liver/gallbladder & Chicken breast/fake blood \\
\hline
ACT & $(500+0)$ & 17/20 (85\%) & 1/20 (5\%) & 1/20 (5\%) \\
ACT & $(500+500)$ & 17/20 (85\%) & 10/20 (50\%) & 4/20 (20\%) \\
ACT$\_PO$ & $(500+0)$ & 17/20 (85\%) & 4/20 (20\%) & 2/20 (10\%) \\
ACT$\_PO$ & $(500+500)$ & 18/20 (90\%) & 13/20 (65\%) & 8/20 (40\%) \\
ACT$\_NA$ & $(500+500)$ & 16/20 (80\%) & 2/20 (10\%) & 2/20 (10\%) \\
ACT & $(1000+0)$ & 19/20 (95\%) & 0/20 (0\%) & 1/20 (5\%) \\
\hline
$\pi_0$ & $(500+0)$ & 12/20 (60\%) & 2/20 (10\%) & 1/20 (5\%) \\
$\pi_0$ & $(500+500)$ & 10/20 (50\%) & 5/20 (25\%) & 1/20 (5\%) \\
$\pi_0$$\_PO$ & $(500+0)$ & 14/20 (70\%) & 1/20 (5\%) & 2/20 (10\%) \\
$\pi_0$$\_PO$ & $(500+500)$ & 15/20 (75\%) & 8/20 (40\%) & 5/20 (25\%) \\
$\pi_0$$\_NA$ & $(500+500)$ & 10/20 (50\%) & 2/20 (10\%) & 3/20 (15\%) \\
$\pi_0$ & $(1000+0)$ & 13/20 (65\%) & 0/20 (0\%) & 0/20 (0\%) \\
\hline
Gr00t-H & $(500+0)$ & 19/20 (95\%) & 8/20 (40\%) & 4/20 (20\%) \\
Gr00t-H & $(500+500)$ & 17/20 (85\%) & 10/20 (50\%) & 7/20 (35\%) \\
Gr00t-H$\_PO$ & $(500+0)$ & 19/20 (95\%) & 9/20 (45\%) & 7/20 (35\%) \\
Gr00t-H$\_PO$ & $(500+500)$ & 19/20 (95\%) & 14/20 (70\%) & 9/20 (45\%) \\
Gr00t-H$\_NA$ & $(500+500)$ & 17/20 (85\%) & 7/20 (35\%) & 4/20 (20\%) \\
Gr00t-H & $(1000+0)$ & 18/20 (90\%) & 5/20 (25\%) & 4/20 (20\%) \\
\hline
\end{tabular}
}
\end{table}

\begin{table}[H]
\centering
\caption*{Full chole task results. Each entry reports successful trials out of 20, with success rate in parentheses.}
\resizebox{\textwidth}{!}{
\begin{tabular}{lcccc}
\hline
Backbone & Training setting & Gallbladder phantom & Normal ex vivo tissue & Abnormal ex vivo tissue \\
\hline
ACT & $(700+0)$ & 16/20 (80\%) & 3/20 (15\%) & 3/20 (15\%) \\
ACT & $(700+700)$ & 17/20 (85\%) & 14/20 (70\%) & 10/20 (50\%) \\
\hline
$\pi_0$ & $(700+0)$ & 10/20 (50\%) & 0/20 (0\%) & 2/20 (10\%) \\
$\pi_0$ & $(700+700)$ & 13/20 (65\%) & 7/20 (35\%) & 4/20 (20\%) \\
\hline
Gr00t-H & $(700+0)$ & 16/20 (80\%) & 4/20 (20\%) & 5/20 (25\%) \\
Gr00t-H & $(700+700)$ & 14/20 (70\%) & 11/20 (55\%) & 10/20 (50\%) \\
\hline
\end{tabular}
}
\end{table}

\end{document}